\documentclass[journal,onecolumn]{IEEEtran}

\IEEEoverridecommandlockouts 

\usepackage{amsmath,amsfonts}
\usepackage{booktabs}
\usepackage{bm}
\usepackage{cite}
\usepackage{color}
\usepackage{siunitx}
\usepackage{graphicx}
\usepackage[nolist]{acronym}
\usepackage{subfig}
\usepackage{float}

\newacro{vs}[VS]{Visual Servoing}
\newacro{ibvs}[IBVS]{Image-Based \ac{vs}}
\newacro{pbvs}[PBVS]{Position-Based \ac{vs}}
\newacro{dvs}[DVS]{Direct \ac{vs}}
\newacro{dof}[DoF]{Degrees of Freedom}
\newacro{nn}[NN]{artificial Neural Network}
\newacro{ml}[ML]{Machine Learning}
\newacro{ssl}[SSL]{Self-Supervised Learning}
\newacro{ooi}[OOI]{Object of Interest}
\newacro{ros}[ROS]{Robot Operating System}
\newacro{il}[IL]{Imitation Learning}
\newacro{dl}[DL]{Deep Learning}
\begin{document}
\title{A Novel Uncalibrated Visual Servoing Controller Baesd on Model-Free Adaptive Control Method with Neural Network}

\author{Haibin Zeng, Yueyong Lyu$^{*}$, Jiaming Qi, Shuangquan Zou, Tanghao Qin, and Wenyu Qin
\thanks{This work was granted by National Key Research and Development Program of China, 2020YFB1506700, and Funded by Science and Technology on Space Intelligent Control Laboratory, No. HTKJ2020KL502014.}
\thanks{All the authors are of School of Astronautics, Harbin Institute of Technology, Harbin, People’s Republic of China\\
}%
}%

\maketitle
\thispagestyle{empty}
\pagestyle{empty}

\begin{abstract}
Nowadays, with the continuous expansion of application scenarios of robotic arms, there are more and more scenarios where nonspecialist come into contact with robotic arms. However, in terms of robotic arm visual servoing, traditional Position-based Visual Servoing (PBVS) requires a lot of calibration work, which is challenging for the nonspecialist to cope with. To cope with this situation, Uncalibrated Image-Based Visual Servoing (UIBVS) frees people from tedious calibration work. This work applied a model-free adaptive control (MFAC) method which means that the parameters of controller are updated in real time, bringing better ability of suppression changes of system and environment. An artificial intelligent neural network is applied in designs of controller and estimator for hand-eye relationship. The neural network is updated with the knowledge of the system input and output information in MFAC method. Inspired by "predictive model" and "receding-horizon" in Model Predictive Control (MPC) method and introducing similar structures into our algorithm, we realizes the uncalibrated visual servoing for both stationary targets and moving trajectories. Simulated experiments with a robotic manipulator will be carried out to validate the proposed algorithm.
\end{abstract}

\section{Introduction}
More and more advances in artificial intelligence and robotics are focusing on autonomy today due to the growing demand for intelligent devicees. By definition, autonomy refers to a device's ability to use data it gathers from a specific situation to perform calculations, calculate probabilities, and make decisions based on logic and the goals they were designed to serve. Unlike an automatic device, which adheres to a set of predetermined rules that specify its behavior regardless of circumstances, it is different from automatic devices.

There are currently several application scenarios, such as manipulating soft materials\cite{soft}, housekeeping\cite{housing}, and intelligent manufacturing\cite{intelligent}, which show that these technologies have a promising future. In these circumstances, it is the responsibility of the devices to provide dependable service that is capable of coping with the real-time situation, and automatic devices that are specialized for a particular circumstance are unable to fulfill the requirements.
For example, the conventional robot arm control uses the kinematic model of the robot arm to design the trajectory of the joint space for the current target\cite{joint_space}. However, when the target changes, the trajectory of the joint space must be redesigned, and it is actually the joint angle signal that serves as the feedback for this control method rather than the system output, which means that it is an open-loop control in the strict sense. When the arm model is modified, the output result will deviate from the anticipated value.

Visual servoing may be roughly classified into two categories: The first type is Position-based Visual Servoing (PBVS), in which a depth camera is used to capture the target position and posture for the feedback signal, therefore forming closed-loop control\cite{PBVS}. The feedback signal acquired by vision sensors must be processed with the data of the system model, and the feedback signal ultimately provided to the controller represents the position and posture of the target relative to the base coordinate system of the robotic manipulator. When the model data are wrong, the estimated position and posture feedback signals may diverge from their actual values when using this method. The alternative type is Image-Based Visual Serving (IBVS). It differs from the PBVS approach in that the IBVS method uses the signal from the vision sensor directly as its feedback signal, hence eliminating numerous unnecessary calculations\cite{compare}.

The estimator for hand-eye relationship is important for the performance and reliability of the system\cite{estimator}. Mnay other work applies a totally online estimator which only uses current input and output data for the hand-eye relationship estimation\cite{Kalman}. Such method can provide the estimator with better ability to adapt to the new circumstance. However, the accuracy of totally online estimator may be far from satisfaction because its data-driven structure and very few data is used for the current estimation\cite{data}. In this work, a neural network-based estimator with a hybrid structure comprising online and offline components is developed. The estimator has been pre-trained using multiple offline data and so provides a reasonably accurate estimate. Once control begins, the estimator will automatically update its parameters using real-time collected data, adapting to changes in the environment and controlled plant. This work's estimator is distinguished by its use of serials history data for updating, its ability to learn from information over time, and its ability to offer a reliable estimate over a period of time, which more closely matches the "receding-horizon" structure in controller.

In this paper the mapping operation from the historical input and error signal to the current input signal is implemented by the proposed controller in the form of a time-varying matrix. Once the formula for the time-varying matrix has been determined, the controller can be constructed. The controller described in this paper was created using the MFAC approach. Initially, a criterion function is constructed. According to the method of traditional MFAC method, the velocity command can then be computed by setting the derivative of the criterion with respect to controller updating components to zero\cite{MFAC}. The controller distinguishes itself by designing an update step size that has been mathematically proven to drive the updating components to their optimal values. In addition, the introduction of a "receding-horizon" structure inspired by the MPC approach distinguishes the proposed controller from conventional ones.

The estimator is calcualted independently with collected data, forming an inner closed loop. While the controller depends on the hand-eye relationship estimation from the inner closed loop for its calculation, creating a larger external closed-loop system. Such constructure with updating algorithm greatly adds to system ability to withstand changes in environments and controlled plant. In addition, the offline trained method provides with a more reliable beginning for both the estimator and controller. And a brief block diagram of the algorithm in this paper can be seen in Fig.1.

\begin{figure}[htbp]
	\centering
	\includegraphics[width=14cm]{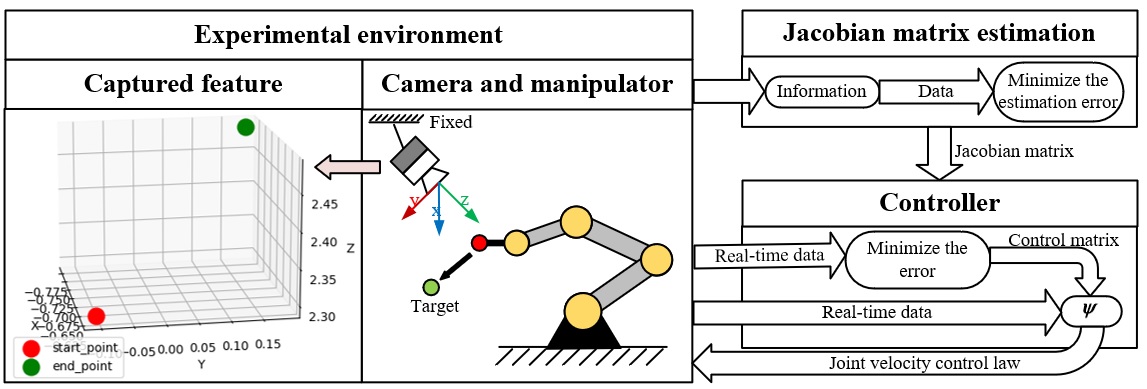}
	\caption{Representation of visual servoing for end-effector position. A red ball is attached to the end of manipulator to make image recognition easier.}
\end{figure}

\section{Related work}

Traditional visual servoing uses vision features abstracted from vision sensor image as feedback signal to achieve a closed-loop system. When it comes to IBVS, the feedback signal directively derive from the image (for example, the coordinates of the center of gravity of the target in the image coordinate system and the area of the target in the image). The features in the feedback signal of the IBVS method are often based on the camera coordinate system. Whereas PBVS uses features that are calculated based on the camera and robot model. The feedbcak signal in PBVS usually comprises features with respect to the robot base coordinate system.

The hand-eye relationship can be roughly divided into two types: the online ones and offline ones. The online ones are usually based on the algorithm like Kalman Filter\cite{KALMAN_FILTER} or BFGS update algorithm\cite{BFGS}. These estimators obtain the estimated values from current and latest historical data. Such structure makes them adaptive to the varying environment. However, their estimation may not be very accurate for the sake of the totally data-driven strategy. The offline ones usually apply neural network (NN)\cite{NN}. In order to make the NN work in all cases, a large number of data must be collected in advance in different situations for the trainning\cite{NN2}. In this case, the work of data collection will take up a lot of time and effort because the control effect totally depends on the offline trained model\cite{NN3}. Wrongly collected data and unsuccessful training will affect the final control effect. 

The controller for uncalibrated visual servoing in other works has been expanded to model predictive control (MPC)\cite{MPC}, model-free adaptive control\cite{MFAC}, and finite-time control\cite{FINITE}. However, few suitable controllers have achieved trajectory tracking under uncalibrated visual servoing conditions. Some optimization based controller like MPC takes a long time to iterate and optimize\cite{MPC2}, which may not be suitable for application scenarios with high real-time requirements. What's more, most the controller mentioned above achieves their velocity command via historical information. Inspired by the observation, we propose a new type of controller represented in the form of a time-varying matrix which maps from the historical information to the current velocity command. By applying the MFAC method, the controller matrix can get the update formula in analytical form, not in optimization form, which means a lot of computational time can be saved.

In this paper, the large amount of data used in trainning is achieved by calculation with the robotic manipulation Denavit-Hartenber (DH) model, such approach can save us a lot of time in collecting data. Our experiments were carried out in the simulation software "CoppeliaSim". It is noted that the model used in the simulation experiment is different from the DH model used in data collection to a certain degree, whcih means that the initial estimation for hand-eye relationship may deviate from true value. We claim that the proposed structure and update algorithm for estimator can effectively drive the estimated value to approach the true value, and can better learn from historical data.

\section{Preliminaries}

This section explains the variable notation used throughout this artical and introduces fundamental concepts of visual servoing.

\subsection{Notation and Nomenclature}

\noindent In this paper, we use standard notation. Lowercase bold letters $\boldsymbol{x}$ indicate column vectors. And uppercase bold letters $\boldsymbol{X}$ indicate matrixes. Time-variant variables are denoted as
$\boldsymbol{x}_{k}$, whose subscript indicates the number of discrete time moment.

In this paper, we focus on the visual servoing for the manipulator’s end-effector without a specific model of the camera or robot. The control effect is realized via the real-time input variable
$\boldsymbol{u}$. To design the corresponding controller, the following premises are necessary.

\begin{itemize}
	\item {The focus of this paper is to design a controller to drive the end-effector of the manipulator moving to the desired position, and the problem of collision or trajectory planning is not the focus of this paper. The posture of the end-effector is not the point of our concern either.  }
	\item {In this paper, the camera is fixed at a specific position and posture, that is, the system is a typical eye-to-hand visual servoing system. Moreover, the depth camera can measure target’s position in camera coordination. The manipulator’s end is bound with an identified object, a red ball, for example, which makes image recognition easier (shown in Fig.1). We define the 3-dimensional position of the end-effector in camera coordination as system state which is written as:  
		\begin{equation}
			{\boldsymbol{y}}=  \left[
			\begin{array}{ccc}
				\boldsymbol{y}_{1} &
				\boldsymbol{y}_{2} &
				\boldsymbol{y}_{3}
			\end{array}
			\right]^{\mathrm{T}}.
	\end{equation}}
	\item {The manipulator is driven by the joint space velocity signal. The control signal is designed in the format of 6-dimensional joint velocity $\dot{\boldsymbol{u}}$. }
\end{itemize}

\subsection{Visual Servoing}
Visual servoing, also known as vision-based robot control and abbreviated VS, is a technique which uses feedback information extracted from a vision sensor to control the motion of a robot\cite{VS}.
In this artical, visual features $\boldsymbol{y}$ serve as feedback. It is directly abstracted from images, becoming part of the closed-loop system.

The features $\boldsymbol{y}$ in this paper does not depend on the camera calibration or Cartesian information because the feedback information and designed trajectory are directly based on the camera coordination and do not need the camera model for calculation.

In this paper, we consider eye-to-hand configurations. Both the robot base and camera base are fixed. In this case, the time derivative of the vision features $\boldsymbol{ \dot y}$ can build relationship with the robot joint space velocity $\boldsymbol{ \dot u}$ with the help of Jacobian matrix $\boldsymbol{J}$ in a local linear format by  $\dot{\boldsymbol{y}} = \boldsymbol{J} \dot{\boldsymbol{u}}$. For the highly nonlinear of visual servoing, the $\boldsymbol{J}$ is time-varying and the estimation of hand-eye relationship in this paper is acually the estimation of the varying Jacobian matrix.

\subsection{Matrix Differentiation}

In this article, the operation for matrix derivation follows the specification in \cite{MATRIX}. 
When it comes to first-order partial derivatives of the transformation from an m-element vector $\boldsymbol{y}$ to an n-element vector $\boldsymbol{x}$, you can get the m×n Jacobian matrix.
\begin{equation}
		\frac{{\partial \boldsymbol{y}}}{{\partial \boldsymbol{x}}} =\left[\begin{array}{ccc}
	\frac{\partial \boldsymbol{y}_{1}}{\partial \boldsymbol{x}_{1}} & \cdots & \frac{\partial \boldsymbol{y}_{1}}{\partial \boldsymbol{x}_{n}} \\
	\vdots & \ddots & \vdots \\
	\frac{\partial \boldsymbol{y}_{m}}{\partial \boldsymbol{x}_{1}} & \cdots & \frac{\partial \boldsymbol{y}_{m}}{\partial \boldsymbol{x}_{n}}
\end{array}\right] 
\end{equation}

We regard scalars as a special case of vectors, and when it comes to first-order partial derivatives of the transformation from a scalar y to an n-element vector $\boldsymbol{x}$, the Jacobian matrix will shrink into row vectors.
\begin{equation}
	\frac{{\partial {y}}}{{\partial \boldsymbol{x}}} =\left[\begin{array}{ccc}
		\frac{\partial {y}}{\partial \boldsymbol{x}_{1}} & \cdots & \frac{\partial {y}}{\partial \boldsymbol{x}_{n}}
	\end{array}\right] 
\end{equation}
which leads to the following proposition:
Let the scalar $\alpha$ be defined by
\begin{equation}
	\alpha = \boldsymbol{y}^{\mathrm {T}} \boldsymbol{A} \boldsymbol{x}
\end{equation}
where $\boldsymbol{y}$ is an m-element vector, $\boldsymbol{x}$ is an n-element vector, $\boldsymbol{A}$ is an m×n matrix, and $\boldsymbol{A}$ is independent of $\boldsymbol{x}$ and $\boldsymbol{y}$, then
\begin{equation}
	\frac{\partial{\alpha}}{{\partial \boldsymbol{x}}} = \boldsymbol{y}^{\mathrm {T}} \boldsymbol{A}
\end{equation}
and
\begin{equation}
	\frac{\partial{\alpha}}{{\partial \boldsymbol{y}}} = \boldsymbol{x}^{\mathrm {T}} \boldsymbol{A}^{\mathrm {T}}
\end{equation}

\section{Method}

In a broad sense, the current control input can be calculated based on current error, historical error, and historical inputs, which is shown in:
\begin{equation}
\boldsymbol{u}(k) = C\left[ {\begin{array}{*{20}{c}}
		{\boldsymbol{e}(k)}& \cdots &{\boldsymbol{e}(k - {n_e})}&{\boldsymbol{u}(k - 1)}& \cdots &{\boldsymbol{u}(k - {n_c})}
\end{array}} \right]
\end{equation}
where ${n_e}$ and ${n_c}$ correspond to the number of errors and inputs considered when calculating the current input. For example, in a traditional uncalibrated Image-based visual servoing controller, the Jacobian matrix that expresses the hand-eye relationship is calculated via online estimation. The controller is then calculated with the estimated Jacobian matrix.
\begin{equation}
\boldsymbol{\dot y} = \boldsymbol{J} \cdot \boldsymbol{\dot u}
\end{equation}
\begin{equation}
	\boldsymbol{\dot u }=  - {K_p}{(\boldsymbol{\hat J})^ + }\boldsymbol{y}
\end{equation}
However, the controller shown from (8) to (9) is a particular case of the controller in Equation.7; that is, this particular case can be expressed using Equation.7 as: 

\begin{equation}
	\boldsymbol{u}(k) =\boldsymbol{u}(k - 1) - {K_p}{(\boldsymbol{\hat J})^ + }\boldsymbol{e}(k)
\end{equation}

By abstracting all the elements when calculating controller in Equation.7, there is: 
\begin{equation}
	\boldsymbol{\xi} (k) = \left[ {\begin{array}{*{20}{c}}
			{\boldsymbol{e}(k)}\\
			\vdots \\
			{\boldsymbol{e}(k - {L_e} + 1)}\\
			{\boldsymbol{u}(k - 1)}\\
			\vdots \\
			{\boldsymbol{u}(k - {L_c})}
	\end{array}} \right] \in {R^{(3{L_e} + 6{L_c}) \times 1}}
\end{equation}
Then Equation.7 can be rewritten as:
\begin{equation}
	\boldsymbol{u}(k) = \boldsymbol{\psi} (k) \cdot \boldsymbol{\xi} (k)
\end{equation}
where $\boldsymbol{\psi} (k)$ is a time-varying matrix calculating the current input via ${L_e}$ history errors and ${L_c}$ history inputs. Here, $\boldsymbol{\psi} (k)$ is divided into $3{L_e} + 6{L_c}$ column vectors (shown in Equation.7). Considering the elements from $\boldsymbol{e}(k)$ to $\boldsymbol{u}(k - {L_c})$ in Equation.1 shows a different importance, there is a gap in value between different columns of the matrix $\boldsymbol{\psi} (k)$. The method of separating matrix into different columns is beneficial to the accuracy of the estimation of matrix $\boldsymbol{\psi} (k)$.
\begin{equation}
	\boldsymbol{\psi} (k) = \left[ {\begin{array}{*{20}{c}}
			{{\boldsymbol{\psi} _1}(k)}& \cdots &{{\boldsymbol{\psi} _{^{3{L_e} + 6{L_c}}}}(k)}
	\end{array}} \right]
\end{equation}
The specific value of the matrix $\boldsymbol{\psi }(k)$ shows the parameter of the controller, which depends on the I/O data of the controller. Since it contains dynamic nonlinearity implicitly, the value can be expressed as a nonlinear function concerning the element vector $\boldsymbol{\xi} (k)$ as shown in the formula.
\begin{equation}
	{\boldsymbol{\psi} _m}(k) = {f_m}\left[ {\boldsymbol{\xi} (k)} \right] \in {R^{6 \times 1}}
\end{equation}

Since the system may generate better performance if the nonlinear characteristic is fully taken into consideration in controller design, the time-varying value of $\boldsymbol{\psi} (k)$ contains certain nonlinear factors. Although the establishment and reference of accurate mathematical models may become a barrier in controller implementation, a balance between performance and practicability is required in this paper.

RBFNN is a forward network with proven strong nonlinear approximation ability and global mapping capacity, whose structure can increase the convergence speed and avoid local minimum, driving us to introduce the RBFNN to the estimation of the controller matrix. What’s more, the simplification of the controller with the local linear matrix $\boldsymbol{\psi} (k)$ and its division will benefit the RBFNN-based approximation method for its simple structure.

\subsection{Controller}

The design of our controller is obtained by minimizing the squared error of the sum of several steps ahead in forecasting, shown in formula (9).
\begin{equation}
    {I_c} =  \sum\limits_{t = 1}^T {\frac{{{\boldsymbol{e}^{\rm{T}}}(k + t)\boldsymbol{e}(k + t)}}{2}} 
\end{equation}
In conventional model-free adaptive control, a similar criterion containning both the system output error as well as input command is designed and by equating the partial derivative of this criterion with respect to the input to zero, the input command can be obtained. It is noted that in this paper we achieve the controller matrix $\boldsymbol{\psi }(k)$ with RBFNN which contains section updated in real time and by calculating the partial derivative of the criterion in (15) with respect to the updated section, the control matrix $\boldsymbol{\psi }(k)$ can therefore be obtained. The principle of the method used in this paper is the same as the MFAC method. The designing of the update law in the proposed method actually corresponds to the coefficient relationship between the two parts of the criterion representing the input and the error in MFAC method. In this case the method we achieve the controller is actually MFAC method. As mentioned before, to obtain ideal input, the partial derivative of the criterion in (15) with respect to the updated section is needed. To get the information, the detail of the controller structure, which will be discussed as follows, should be built to acquire an ideal prediction of control law using the input and output data of the system. In which way the optimization procedure of Equation.15 can be acquired.


The structure of RBFNN is used to approximate the function ${f_m}\left(  \bullet  \right)$ in the Equation.14. It is evident that the input of these RFBNNs is the information vector $\boldsymbol{\xi} (k)$ and the output is the corresponding ${m^{th}}$ column of the controller matrix $\boldsymbol{\psi} (k)$. 

The value of the ${n^{th}}$ hidden neuron of the ${m^{th}}$ RBFNN is
\begin{equation}
	\boldsymbol{h_{mn}}(k) =  {e^{ - \frac{{\left\| {\boldsymbol{\xi} (k) - {\boldsymbol{c}_n}} \right\|}}{{\boldsymbol{\gamma} _n^2}}}},{\rm{    }}n = 1, \ldots ,{N_h}.
\end{equation}
Where ${N_h}$ is the number of hidden neuron of the ${m^{th}}$ RBFNN. ${\boldsymbol{\gamma} _n}$ is the radius of the ${n^{th}}$ hidden neuron, and ${c_n}$ denotes the center of the ${n^{th}}$ hidden neuron. The output of the ${m^{th}}$ neural network is
\begin{equation}
	{\boldsymbol{\psi} _m}(k)  = {f_m}\left[ {\boldsymbol{\xi} (k)} \right] = \sum\limits_{n = 1}^{{N_h}} {{\boldsymbol{w}_{mn}}(k){\boldsymbol{h}_{mn}}(k)} 	= {\boldsymbol{W}_m} \cdot {\boldsymbol{h}_m}(k){\rm{  }},m = 1, \ldots ,3{L_e} + 6{L_c}
\end{equation}
where ${\boldsymbol{W}_m}$ is the weight matrix of the $m^{th}$ RBFNN and ${\boldsymbol{w}_{mn}}$ is the $n^{th}$ column of ${\boldsymbol{W}_m}$. Since the controller is built using RBFNN, the parameter adjustment should be given priority consideration for the whole system’s convergency performance and robustness. An effective and practical update method will be discussed, which can update the full connect layer of RBFNN, thus improving the robustness of the system against changes in the environment and reducing the impact of insufficient offline training or erroneous in training data set.

Assumption 3.1: There exists an ideal weight matrix ${\boldsymbol{W}_m}^ * $ for each RBFNN, and it can be obtained that 
${\boldsymbol{\psi} _m}^ * (k) = {f_m}^ * \left[ {\boldsymbol{\xi} (k)} \right] = \sum\limits_{n = 1}^{{N_h}} {{\boldsymbol{w}^ * }_{mn}(k){\boldsymbol{h}^ * }_{mn}(k)}  = {\boldsymbol{W}^ * }_m \cdot {\boldsymbol{h}^ * }_m(k)$.
The ideal control matrix is ${\boldsymbol{\psi} ^ * }(k) = \left[ {\begin{array}{*{20}{c}}
		{{\boldsymbol{\psi} _1}^ * (k)}& \cdots &{\boldsymbol{\psi} _{3{L_e} + 6{L_c}}^ * (k)}
\end{array}} \right]$. And the input calculated via ${\boldsymbol{\psi} ^ * }(k)$ is supposed to drive the system error in several steps ahead converge to zero.

By calculating the gradient of ${I_c}$ with respect to ${\boldsymbol{\hat W}_{mn}}(k)$, it is obvious that in order to get specific numeric of $\frac{{\partial {I_c}}}{{\partial {\boldsymbol{\hat W}_{mn}}(k)}}$, the estimation of $\frac{{\partial \boldsymbol{y}(k + 1)}}{{\partial \boldsymbol{u}(k)}}$ is needed to be carried out first. In the following section, there will be a detailed description of the estimation of $\frac{{\partial \boldsymbol{y}(k + 1)}}{{\partial \boldsymbol{u}(k)}}$. The specific calculation formula of $\frac{{\partial {I_c}}}{{\partial {\boldsymbol{\hat W}_{mn}}(k)}}$ is:
\begin{equation}
	\begin{aligned}
		\frac{{\partial {I_c}}}{{\partial {\boldsymbol{\hat W}_{mn}}(k)}} =& \frac{{\partial \left( {\sum\limits_{t = 1}^T {\frac{{{\boldsymbol{e}^{\rm{T}}}(k + t)\boldsymbol{e}(k + t)}}{2}} } \right)}}{{\partial {\boldsymbol{e}}(k + t)}}  \frac{{\partial {\boldsymbol{e}}(k + t)}}{{\partial \boldsymbol{u}(k + t - 1)}}\frac{{\partial \boldsymbol{u}(k + t - 1)}}{{\partial {\boldsymbol{\hat \psi }_m}(k + t - 1)}}\frac{{\partial {\boldsymbol{\hat \psi }_m}(k + t - 1)}}{{\partial {\boldsymbol{\hat w}_{mn}}(k + t - 1)}}\\
		\frac{{\partial {I_c}}}{{\partial {\boldsymbol{\hat W}_{mn}}(k)}} =&  - \sum\limits_{t = 1}^T {{\boldsymbol{e}^{\rm{T}}}(k + t)\frac{{\partial \boldsymbol{y}(k + 1)}}{{\partial \boldsymbol{u}(k + t - 1)}}{\boldsymbol{\bar \xi }_m}(k + t - 1){\boldsymbol{h}_{mn}}(k + t - 1)} .
	\end{aligned}
\end{equation}
Then the update law can be expressed with steepest descent algorithm as follows.
\begin{equation}
	{\boldsymbol{\hat W}_{mn}}(k{\rm{ + }}1){\rm{ = }}  \boldsymbol{\hat W_{mn}}(k){\rm{ + }} \left[ \alpha \sum\limits_{t = 1}^T {{\boldsymbol{e}^{\rm{T}}}(k + t)\frac{{\partial \boldsymbol{y}(k + 1)}}{{\partial \boldsymbol{u}(k + t - 1)}}{\boldsymbol{\bar \xi }_m}(k + t - 1){\boldsymbol{h}_{mn}}(k + t - 1)} \right] ^{\rm{T}}
\end{equation}
where $\alpha $ is a negative scale according to the step size. Although the update algorithm is supposed to lead the criterion ${I_c} =  \sum\limits_{t = 1}^T {\frac{{{\boldsymbol{e}^{\rm{T}}}(k + t)\boldsymbol{e}(k + t)}}{2}} $ to zero, the step size needs to be designed with the aim of convergency of the whole system.

Here we define the ${m^{th}}$ column of the current error of the ${n^{th}}$ weights matrix as: 
\begin{equation}
	{\boldsymbol{\tilde W}_{mn}}(k){\rm{ = }}  \boldsymbol{W^ * }_{mn} - {\boldsymbol{\hat W}_{mn}}(k).
\end{equation}
We abbreviate ${\boldsymbol{W}}^ * $ as ${\boldsymbol{W}} $ in the following content. Introduce (19) into (20), and we can get
\begin{equation}
	\begin{aligned}
		{\boldsymbol{\tilde W}_{mn}}(k{\rm{ + }}1)= & {\boldsymbol{W}_{mn}} - {\boldsymbol{\hat W}_{mn}}(k{\rm{ + }}1)\\
		=  & {\boldsymbol{W}_{mn}} -   {\boldsymbol{\hat W}_{mn}}(k){\rm{ - }} \left[ \alpha \sum\limits_{t = 1}^T {{\boldsymbol{e}^{\rm{T}}}(k + t)\frac{{\partial \boldsymbol{y}(k + 1)}}{{\partial \boldsymbol{u}(k + t - 1)}}{\boldsymbol{\bar \xi }_m}(k + t - 1){\boldsymbol{h}_{mn}}(k + t - 1)}\right] ^{\rm{T}} \\
		=  & {\boldsymbol{\tilde W}_{mn}}(k){\rm{ - }} \left[ \alpha \sum\limits_{t = 1}^T {{\boldsymbol{e}^{\rm{T}}}(k + t)\frac{{\partial \boldsymbol{y}(k + 1)}}{{\partial \boldsymbol{u}(k + t - 1)}}{\boldsymbol{\bar \xi }_m}(k + t - 1){\boldsymbol{h}_{mn}}(k + t - 1)}\right] ^{\rm{T}}.
	\end{aligned}
\end{equation}
To design a reasonable step size of update algorithm that drives ${\hat W_{mn}}(k)$ to ${W^ * }_{mn}$, a corresponding cost function is set as
\begin{equation}
	\begin{aligned}
		{V_1}(k + 1) =  & {\sum\limits_m {\sum\limits_n {\left\| {{\boldsymbol{\tilde W}_{mn}}(k + 1)} \right\|} } ^2}\\
		=  & {\sum\limits_m {\sum\limits_n {\left\| {{\boldsymbol{\tilde W}_{mn}}(k){\rm{ - }} \left[ \alpha \sum\limits_{t = 1}^T {{\boldsymbol{e}^{\rm{T}}}(k + t)\frac{{\partial \boldsymbol{y}(k + 1)}}{{\partial \boldsymbol{u}(k + t - 1)}}{\boldsymbol{\bar \xi }_m}(k + t - 1){\boldsymbol{h}_{mn}}(k + t - 1)}\right] ^{\rm{T}} } \right\|} } ^2}\\
		=  & {\sum\limits_m {\sum\limits_n {\left\| {{\boldsymbol{\tilde W}_{mn}}(k)} \right\|} } ^2} + \sum\limits_m {\sum\limits_n {{\alpha ^2}} } {\left\| {\sum\limits_{t = 1}^T {{\boldsymbol{e}^{\rm{T}}}(k + t)\frac{{\partial \boldsymbol{y}(k + 1)}}{{\partial \boldsymbol{u}(k + t - 1)}}{\boldsymbol{\bar \xi }_m}(k + t - 1){\boldsymbol{h}_{mn}}(k + t - 1)} } \right\|^2}\\
		&  - \sum\limits_m {\sum\limits_n {2\alpha \sum\limits_{t = 1}^T {\left[ {{\boldsymbol{e}^{\rm{T}}}(k + t)\frac{{\partial \boldsymbol{y}(k + 1)}}{{\partial \boldsymbol{u}(k + t - 1)}}{\boldsymbol{\bar \xi }_m}(k + t - 1){\boldsymbol{h}_{mn}}(k + t - 1)} \right]} } } {\boldsymbol{\tilde W}_{mn}}(k).
	\end{aligned}
\end{equation}
The step size is expressed as
\begin{equation}
	\alpha  =   \frac{{ - k\sum\limits_{t = 1}^T {{{\left\| {{\boldsymbol{e}^{\rm{T}}}(k + t)} \right\|}^2}} }}{{\sum\limits_m {\sum\limits_n {{{\left[ {\sum\limits_{t = 1}^T {{\boldsymbol{e}^{\rm{T}}}(k + t)\frac{{\partial \boldsymbol{y}(k + t)}}{{\partial \boldsymbol{u}(k + t - 1)}}{\boldsymbol{\bar \xi }_m}(k + t - 1){\boldsymbol{h}_{mn}}(k + t - 1)} } \right]}^2}} } }}.
\end{equation}
From ${\boldsymbol{\psi} _m}(k) = {f_m}\left[ {\boldsymbol{\xi} (k)} \right] = \sum\limits_{n = 1}^{{N_h}} {{\boldsymbol{w}_{mn}}(k){\boldsymbol{h}_{mn}}(k)} {\rm{  }},m = 1, \ldots ,3{L_e} + 6{L_c}$ and $\boldsymbol{u}(k) = \boldsymbol{\psi} (k) \cdot \boldsymbol{\xi} (k)$, we can get that
\begin{equation}
	{\boldsymbol{\tilde \psi} _m}(k) = \sum\limits_{n = 1}^{{N_h}} {{\boldsymbol{\tilde w}_{mn}}(k){\boldsymbol{h}_{mn}}(k)} ,m = 1, \ldots ,3{L_e} + 6{L_c}
\end{equation}
\begin{equation}
	\boldsymbol{\tilde u}(k) = \boldsymbol{\tilde \psi} (k) \cdot \boldsymbol{\xi} (k)
\end{equation}
By introducing (23),(24) and(25) into (22), the variation of Lyapunov function in discrete form is expressed as
\begin{equation}
	\begin{aligned}
		{V_1}(k + 1) - {V_1}(k)
		&  =    - \sum\limits_m {\sum\limits_n {2\alpha \sum\limits_{t = 1}^T {\left[ {{\boldsymbol{e}^{\rm{T}}}(k + t)\frac{{\partial \boldsymbol{y}(k + 1)}}{{\partial \boldsymbol{u}(k + t - 1)}}{\boldsymbol{\bar \xi }_m}(k + t - 1){\boldsymbol{h}_{mn}}(k + t - 1)} \right]} } } {\boldsymbol{\tilde W}_{mn}}(k)  \\
		&  + \sum\limits_m {\sum\limits_n {{\alpha ^2}{{\left[ {\sum\limits_{t = 1}^T {{\boldsymbol{e}^{\rm{T}}}(k + t)\frac{{\partial \boldsymbol{y}(k + 1)}}{{\partial \boldsymbol{u}(k + t - 1)}}{\boldsymbol{\bar \xi }_m}(k + t - 1){\boldsymbol{h}_{mn}}(k + t - 1)} } \right]}^2}} } \\
		&  =   \sum\limits_m {\sum\limits_n {\frac{{2k\sum\limits_{t = 1}^T {{{\left\| {{\boldsymbol{e}^{\rm{T}}}(k + t)} \right\|}^2}} \sum\limits_{t = 1}^T {\left[ {{\boldsymbol{e}^{\rm{T}}}(k + t)\frac{{\partial \boldsymbol{y}(k + 1)}}{{\partial \boldsymbol{u}(k + t - 1)}}{\boldsymbol{\bar \xi }_m}(k + t - 1){\boldsymbol{h}_{mn}}(k + t - 1)} \right]{\boldsymbol{\tilde W}_{mn}}(k)} }}{{\sum\limits_m {\sum\limits_n {{{\left[ {\sum\limits_{t = 1}^T {{\boldsymbol{e}^{\rm{T}}}(k + t)\frac{{\partial \boldsymbol{y}(k + t)}}{{\partial \boldsymbol{u}(k + t - 1)}}{\boldsymbol{\bar \xi }_m}(k + t - 1){\boldsymbol{h}_{mn}}(k + t - 1)} } \right]}^2}} } }}}   } \\
		&    + \sum\limits_m {\sum\limits_n {\frac{{{{\left[ { - k\sum\limits_{t = 1}^T {{{\left\| {{\boldsymbol{e}^{\rm{T}}}(k + t)} \right\|}^2}} } \right]}^2}{{\left[ {\sum\limits_{t = 1}^T {{\boldsymbol{e}^{\rm{T}}}(k + t)\frac{{\partial \boldsymbol{y}(k + 1)}}{{\partial \boldsymbol{u}(k + t - 1)}}{\boldsymbol{\bar \xi }_m}(k + t - 1){\boldsymbol{h}_{mn}}(k + t - 1)} } \right]}^2}}}{{{{\left\{ {\sum\limits_m {\sum\limits_n {{{\left[ {\sum\limits_{t = 1}^T {{\boldsymbol{e}^{\rm{T}}}(k + t)\frac{{\partial \boldsymbol{y}(k + t)}}{{\partial \boldsymbol{u}(k + t - 1)}}{\boldsymbol{\bar \xi }_m}(k + t - 1){\boldsymbol{h}_{mn}}(k + t - 1)} } \right]}^2}} } } \right\}}^2}}}} } \\
		&  = \frac{{2k\sum\limits_{t = 1}^T {{{\left\| {{\boldsymbol{e}^{\rm{T}}}(k + t)} \right\|}^2}\sum\limits_m {\sum\limits_{t = 1}^T {\left[ {{\boldsymbol{e}^{\rm{T}}}(k + t)\frac{{\partial \boldsymbol{y}(k + 1)}}{{\partial \boldsymbol{u}(k + t - 1)}}{\boldsymbol{\bar \xi }_m}(k + t - 1){\boldsymbol{\tilde \psi }_m}(k + t - 1)} \right]} }   } }}{{\sum\limits_m {\sum\limits_n {{{\left[ {\sum\limits_{t = 1}^T {{\boldsymbol{e}^{\rm{T}}}(k + t)\frac{{\partial \boldsymbol{y}(k + t)}}{{\partial \boldsymbol{u}(k + t - 1)}}{\boldsymbol{\bar \xi }_m}(k + t - 1){\boldsymbol{h}_{mn}}(k + t - 1)} } \right]}^2}} } }}\\
		&    + \frac{{{{\left[ { - k\sum\limits_{t = 1}^T {{{\left\| {{\boldsymbol{e}^{\rm{T}}}(k + t)} \right\|}^2}} } \right]}^2}}}{{\sum\limits_m {\sum\limits_n {{{\left[ {\sum\limits_{t = 1}^T {{\boldsymbol{e}^{\rm{T}}}(k + t)\frac{{\partial \boldsymbol{y}(k + t)}}{{\partial \boldsymbol{u}(k + t - 1)}}{\boldsymbol{\bar \xi }_m}(k + t - 1){\boldsymbol{h}_{mn}}(k + t - 1)} } \right]}^2}} } }}\\
		&  =   \frac{{2k{{\left[ {\sum\limits_{t = 1}^T {{{\left\| {{\boldsymbol{e}^{\rm{T}}}(k + t)} \right\|}^2}} } \right]}^2}}}{{\sum\limits_m {\sum\limits_n {{{\left[ {\sum\limits_{t = 1}^T {{\boldsymbol{e}^{\rm{T}}}(k + t)\frac{{\partial \boldsymbol{y}(k + t)}}{{\partial \boldsymbol{u}(k + t - 1)}}{\boldsymbol{\bar \xi }_m}(k + t - 1){\boldsymbol{h}_{mn}}(k + t - 1)} } \right]}^2}} } }}\\
		&    + \frac{{{k^2}{{\left[ {\sum\limits_{t = 1}^T {{{\left\| {{\boldsymbol{e}^{\rm{T}}}(k + t)} \right\|}^2}} } \right]}^2}}}{{\sum\limits_m {\sum\limits_n {{{\left[ {\sum\limits_{t = 1}^T {{\boldsymbol{e}^{\rm{T}}}(k + t)\frac{{\partial \boldsymbol{y}(k + t)}}{{\partial \boldsymbol{u}(k + t - 1)}}{\boldsymbol{\bar \xi }_m}(k + t - 1){\boldsymbol{h}_{mn}}(k + t - 1)} } \right]}^2}} } }}\\
		&  =   \frac{{(2k + {k^2}){{\left[ {\sum\limits_{t = 1}^T {{{\left\| {{\boldsymbol{e}^{\rm{T}}}(k + t)} \right\|}^2}} } \right]}^2}}}{{\sum\limits_m {\sum\limits_n {{{\left[ {\sum\limits_{t = 1}^T {{\boldsymbol{e}^{\rm{T}}}(k + t)\frac{{\partial \boldsymbol{y}(k + t)}}{{\partial \boldsymbol{u}(k + t - 1)}}{\boldsymbol{\bar \xi }_m}(k + t - 1){\boldsymbol{h}_{mn}}(k + t - 1)} } \right]}^2}} } }}
	\end{aligned}
\end{equation}
In order to drive the full connect matrix to approach ideal matrix, ${V_1}(k + 1) - {V_1}(k)$ should be negative, and the negative number ranges from -2 to 0. 

The controlled plant is unknown, and the specific value of $\frac{{\partial \boldsymbol{e}(k + 1)}}{{\partial \boldsymbol{u}(k)}}$ in (23) and (26) still needs algorithm for estimation. The accuracy of controlled plant estimation greatly affects whole system’s robustness and performance. Once a fast and reliable estimation algorithm is established, many control methods such one step ahead prediction or backstepping can be constructed based on it. In the following section, a neural network is built to mimic the relationship between $\Delta \boldsymbol{e}(k + 1)$ and $\Delta \boldsymbol{u}(k)$, which completes the prediction of $\frac{{\partial \boldsymbol{e}(k + 1)}}{{\partial \boldsymbol{u}(k)}}$.

\subsection{Estimator for Controlled Plant}
The hand-eye relationship is highly coupled and non-linear. An efficient method is to expressed the relationship in local linear format using Jacobian matrix like
\begin{equation}
	\boldsymbol{\dot y} = \boldsymbol{J} \cdot \boldsymbol{\dot u}.
\end{equation}
Due to the characteristic of hand-eye relationship, the Jacobian matrix is time-varying. It is a proved efficient method to estimate the hand-eye relationship by calculating the time-varying Jacobian matrix through algorithm related to Kalman Filter or Neural Network. Once the $\boldsymbol{J}$ in (27) is calculated, the needed prediction of $\frac{{\partial \boldsymbol{y}(k + 1)}}{{\partial \boldsymbol{u}(k)}}$ can be found. And the estimation algorithm for $\boldsymbol{J}$ is the main work of this section.

In many traditional methods, the estimation of $J$ is carried out in purely online or offline way. The purely online method is usually applied with Kalman Filter, which possesses good stability but shows insufficient convergency performance. Because only the input and output data is involved when calculating Jacobian matrix. There is hardly any information about controlled plant. While, another kind of method for approximation in purely offline way usually realizes its function via neural network, for instance. When the environment is as expected, the collected data used in pre-training procedure benefits whole system’s performance. This kind of algorithm can perform well in certain situation, that is, they are susceptible to environment changes because training set data is collected under certain conditions which may change in actual operation. In this section, an effective algorithm with both online part and offline part is proposed, acquiring a balance between the general applicability and performance of estimation.

In offline phrase, a RBFNN is constructed to map from current value of joint space to current Jacobian matrix. The training data set is collected on the premise that transfer matrix from the base coordinate system of robot arm to the body coordinate matrix of the depth camera is a fixed value. The Jacobian matrix is divided into six columns (shown in Equation.28), estimated by six independent neural networks. There is a difference in the size of the values. The employment of different learning rate accelerates the convergency of offline training phrase. 
\begin{equation}
	\begin{aligned}
		\boldsymbol{J}{\rm{ = }} & \left[ {\begin{array}{*{20}{c}}
				{\frac{{\partial \boldsymbol{x_1}}}{{\partial \boldsymbol{r_1}}}}& \cdots &{\frac{{\partial \boldsymbol{x_1}}}{{\partial \boldsymbol{r_6}}}}\\
				\vdots & \ddots & \vdots \\
				{\frac{{\partial \boldsymbol{x_3}}}{{\partial \boldsymbol{r_1}}}}& \cdots &{\frac{{\partial \boldsymbol{x_3}}}{{\partial \boldsymbol{r_6}}}}
		\end{array}} \right]\\
		\boldsymbol{J} =  & \left[ {\begin{array}{*{20}{c}}
				{\boldsymbol{J_1}}&{\boldsymbol{J_2}}&{\boldsymbol{J_3}}&{\boldsymbol{J_4}}&{\boldsymbol{J_5}}&{\boldsymbol{J_6}}
		\end{array}} \right]
	\end{aligned}
\end{equation}

The hidden neuron is written as
\begin{equation}
	{\boldsymbol{\theta} _{{\rm{it}}}} = {{\rm{e}}^{{\rm{ - }}\frac{{||\boldsymbol{r} - {\boldsymbol{u}_{{\rm{it}}}}||}}{{{\boldsymbol{\delta} _{{\rm{it}}}}}}{^2}}}.
\end{equation}
Here, ${\boldsymbol{\delta} _{{\rm{it}}}}$ denotes the radius of the ${\rm{t}^{th}}$ hidden neuron. ${u_{{\rm{it}}}}$ denotes the center of the ${\rm{t}^{th}}$ hidden neuron. And the output of the ${\rm{i}^{th}}$ neural network is expressed as
\begin{equation}
	{\boldsymbol{J}_{\rm{i}}} = {\boldsymbol{W}_{\rm{i}}}{\boldsymbol{\theta} _{\rm{i}}}(\boldsymbol{r}){\rm{i}} = 1,...,6
\end{equation}

To design the update law for estimation neural network, a suitable criterion is required. It is obvious that the estimated value is the change of characteristic of end-effector over a period of time. Then, the corresponding criterion for estimation accuracy is
\begin{equation}
	\begin{aligned}
		Q\left( k \right){\rm{ = }} & \sum\limits_{t = 1}^T {{{\left[ {\boldsymbol{y}\left( {k + 1 - t} \right) -\boldsymbol{ \hat y}\left( {k + 1 - t} \right)} \right]}^{\rm{T}}}\left[ {\boldsymbol{y}\left( {k + 1 - t} \right) - \boldsymbol{\hat y}\left( {k + 1 - t} \right)} \right]} \\
		{\rm{ = }} & \sum\limits_{t = 1}^T {{{\left[ {\boldsymbol{\Delta y}\left( {k + 1 - t} \right) - \boldsymbol{\Delta \hat y}\left( {k + 1 - t} \right)} \right]}^{\rm{T}}}\left[ {\boldsymbol{\Delta y}\left( {k + 1 - t} \right) - \boldsymbol{\Delta \hat y}\left( {k + 1 - t} \right)} \right]} 
	\end{aligned}
\end{equation}
By calculating the relationship between $Q\left( {k + 1} \right)$ and ${{\boldsymbol{\hat w}}_{mn}}\left( {k + 1} \right)$ , the updating gradient direction is obtained as
\begin{equation}
	\begin{aligned}
		\frac{{\partial Q\left( k \right)}}{{\partial {{{\boldsymbol{\hat w}}}_{mn}}\left( k \right)}} =  & \sum\limits_{t = 1}^T {\frac{{\partial Q\left( k \right)}}{{\partial \boldsymbol{\Delta \hat y}\left( {k + 1 - t} \right)}}\frac{{\partial \boldsymbol{\Delta \hat y}\left( {k + 1 - t} \right)}}{{\partial \boldsymbol{J_m}\left( {k + 1 - t} \right)}}\frac{{\partial \boldsymbol{J_m}\left( {k + 1 - t} \right)}}{{\partial {{{\boldsymbol{\hat w}}}_{mn}}\left( k \right)}}} \\
		\frac{{\partial Q\left( k \right)}}{{\partial {{{\boldsymbol{\hat w}}}_{mn}}\left( k \right)}} =  & \sum\limits_{t = 1}^T { - {{\left[ {\boldsymbol{\Delta y}\left( {k + 1 - t} \right) - \boldsymbol{\Delta \hat y}\left( {k + 1 - t} \right)} \right]}^{\rm{T}}}\boldsymbol{\Delta {u_m}}\left( {k + 1 - t} \right)\frac{{\partial \boldsymbol{J_m}\left( {k + 1 - t} \right)}}{{\partial {{{\boldsymbol{\hat w}}}_{mn}}\left( k \right)}}} \\
		\frac{{\partial Q\left( k \right)}}{{\partial {{{\boldsymbol{\hat w}}}_{mn}}\left( k \right)}} =  & \sum\limits_{t = 1}^T { - {{\left[ {\boldsymbol{\Delta y}\left( {k + 1 - t} \right) - \boldsymbol{\Delta \hat y}\left( {k + 1 - t} \right)} \right]}^{\rm{T}}}\boldsymbol{\Delta {u_m}}\left( {k + 1 - t} \right){\boldsymbol{\theta} _{mn}}(k + 1 - t)} 
	\end{aligned}
\end{equation}
Here the updating law for full connect layer of estimation neural network is expressed as 
\begin{equation}
\boldsymbol{\hat W_{mn}}(k{\rm{ + }}1){\rm{ = }}  \boldsymbol{\hat W_{mn}}(k){\rm{ - }}{\alpha _2}\sum\limits_{t = 1}^T {{\left[ {\boldsymbol{\Delta y}\left( {k + 1 - t} \right) -\boldsymbol{ \Delta \hat y}\left( {k + 1 - t} \right)} \right]}\boldsymbol{\Delta {u_m}}\left( {k + 1 - t} \right)\boldsymbol{\theta _{mn}}(k + 1 - t)} .
\end{equation}

Assumption 3.2: There exists an ideal full connect layer parameter ${\boldsymbol{W}_m}$ for the ${m^{th}}$ RBFNN, and it can be obtained that ${\boldsymbol{J}_m}^ * (k) = {g_m}^ * \left[ {\boldsymbol{u}(k)} \right] = \sum\limits_{n = 1}^{{N_h}} {\boldsymbol{w}_{mn}^ * (k)\boldsymbol{\theta} _{mn}^{}(k)}  = \boldsymbol{W}_m^ *  \cdot \boldsymbol{h}_m^{}(k)$. The ideal Jacobian matrix is ${\boldsymbol{J}^ * }(k) = \left[ {\begin{array}{*{20}{c}}
		{\boldsymbol{J_1}^ * (k)} \cdots {\boldsymbol{J_6}^ * (k)}
\end{array}} \right]$. And the error of estimated steps ahead output calculated via ${\boldsymbol{J}^ * }(k)$ is supposed to converge to zero for the sake of updating law in (33).

The updating law is supposed to drive the full connect matrix ${\boldsymbol{\hat W}_{mn}}(k{\rm{ + }}1)$ to an ideal one ${\boldsymbol{W}_{mn}}$. To evaluate the erroneous of full connect layer, here we define the deviation of full connect matrix
\begin{equation}
	\begin{aligned}
		{\boldsymbol{\tilde W}_{mn}}(k{\rm{ + }}1){\rm{ = }} & {\boldsymbol{W}_{mn}} - {\boldsymbol{\hat W}_{mn}}(k{\rm{ + }}1)\\
		=  & {\boldsymbol{\tilde W}_{mn}}(k){\rm{ + }}{\alpha _2}\sum\limits_{t = 1}^T {{{\left[ {\boldsymbol{\Delta y}\left( {k + 1 - t} \right) - \boldsymbol{\Delta \hat y}\left( {k + 1 - t} \right)} \right]}}\boldsymbol{\Delta u}_m\left( {k + 1 - t} \right){\boldsymbol{\theta} _{mn}}(k + 1 - t)} .
	\end{aligned}
\end{equation}
where the number ${\alpha _2}$ is a negative scale who acts as the step size of updating law. In (34), the full connect layer is moving in the direction of a negative gradient. However, a suitable step size needs to be designed with convergency proof. A Lyapunov function related to deviation of full connect layer is expressed as
\begin{equation}
	\begin{aligned}
		{V_2}(k + 1) =  & {\sum\limits_m {\sum\limits_n {\left\| {{\boldsymbol{\tilde W}_{mn}}(k + 1)} \right\|} } ^2}\\
		=  & {\sum\limits_m {\sum\limits_n {\left\| {{\boldsymbol{\tilde W}_{mn}}(k){\rm{ + }}{\alpha _2}\sum\limits_{t = 1}^T {{{\left[ {\boldsymbol{\Delta y}\left( {k + 1 - t} \right) - \boldsymbol{\Delta \hat y}\left( {k + 1 - t} \right)} \right]}}\boldsymbol{\Delta u}_m\left( {k + 1 - t} \right){\boldsymbol{\theta} _{mn}}(k + 1 - t)} } \right\|} } ^2}\\
		=  & {V_2}(k) + \sum\limits_m {\sum\limits_n {2{\alpha _2}\sum\limits_{t = 1}^T {{{\left[ {\boldsymbol{\Delta y}\left( {k + 1 - t} \right) - \boldsymbol{\Delta \hat y}\left( {k + 1 - t} \right)} \right]}}\boldsymbol{\Delta u}_m\left( {k + 1 - t} \right){\boldsymbol{\theta} _{mn}}(k + 1 - t)} } } \boldsymbol{\tilde W}_{mn}(k)\\
		&  + \alpha _2^2\sum\limits_m {\sum\limits_n {{{\left\| {\sum\limits_{t = 1}^T {{{\left[ {\boldsymbol{\Delta y}\left( {k + 1 - t} \right) - \boldsymbol{\Delta \hat y}\left( {k + 1 - t} \right)} \right]}^{\rm{T}}}\boldsymbol{\Delta u}_m\left( {k + 1 - t} \right){\boldsymbol{\theta} _{mn}}(k + 1 - t)} } \right\|}^2}} } .
	\end{aligned}
\end{equation}
And the step size in (33) is designed as
\begin{equation}
	{\alpha _2} =  \frac{{ - {k_2}}}{{\sum\limits_{t = 0}^T {\sum\limits_m {\boldsymbol{\Delta u}_m^2\left( {k - t} \right)\boldsymbol{\theta} _m^2\left( {k - t} \right)} } }}.
\end{equation}
From ${\boldsymbol{J}_{\rm{i}}} = {\boldsymbol{W}_{\rm{i}}}{\boldsymbol{\theta} _{\rm{i}}}(r){\rm{i}} = 1,...,6$ and $\boldsymbol{\dot x} = \boldsymbol{J} \cdot \boldsymbol{\dot u}$, It can be obtained that 
\begin{equation}
	\boldsymbol{\tilde J}_m(k) = \sum\limits_{n = 1}^{{N_h}} {\boldsymbol{\tilde w}_{mn}^{}(k)\boldsymbol{\theta} _{mn}^{}(k)}  = \boldsymbol{\tilde W}_m^{} \cdot \boldsymbol{h}_m^{}(k)
\end{equation}
\begin{equation}
	\boldsymbol{\Delta \tilde y} = \boldsymbol{\tilde J} \cdot \boldsymbol{\Delta u}
\end{equation}
By introducing (36),(37) and(38) into (35), the variation of Lyapunov function in discrete form is expressed as
\begin{equation}
	\begin{aligned}
		{V_2}(k + 1) - {V_2}(k) =  & \sum\limits_m {\sum\limits_n {\frac{{ - 2{k_2}\sum\limits_{t = 1}^T {{{\left[ {\boldsymbol{\Delta y}\left( {k + 1 - t} \right) - \boldsymbol{\Delta \hat y}\left( {k + 1 - t} \right)} \right]}^{\rm{T}}}\boldsymbol{\Delta u}_m\left( {k + 1 - t} \right){\boldsymbol{\theta} _{mn}}(k + 1 - t)} }}{{\sum\limits_{t = 0}^T {\sum\limits_m {\boldsymbol{\Delta u}_m^2\left( {k - t} \right)\boldsymbol{\theta} _m^2\left( {k - t} \right)} } }}} } \boldsymbol{\tilde W}_{mn}(k)\\
		&  + \frac{{k_2^2\sum\limits_m {\sum\limits_n {{{\left\| {\sum\limits_{t = 1}^T {{{\left[ {\boldsymbol{\Delta y}\left( {k + 1 - t} \right) - \boldsymbol{\Delta \hat y}\left( {k + 1 - t} \right)} \right]}^{\rm{T}}}\boldsymbol{\Delta u}_m\left( {k + 1 - t} \right){\boldsymbol{\theta} _{mn}}(k + 1 - t)} } \right\|}^2}} } }}{{{{\left[ {\sum\limits_{t = 0}^T {\sum\limits_m {\boldsymbol{\Delta u}_m^2\left( {k - t} \right)\boldsymbol{\theta} _m^2\left( {k - t} \right)} } } \right]}^2}}}\\
		\le  & \sum\limits_m {\sum\limits_n {\frac{{ - 2{k_2}\sum\limits_{t = 1}^T {{{\left[ {\boldsymbol{\Delta y}\left( {k + 1 - t} \right) - \boldsymbol{\Delta \hat y}\left( {k + 1 - t} \right)} \right]}^{\rm{T}}}\boldsymbol{\Delta u}_m\left( {k + 1 - t} \right){\boldsymbol{\theta} _{mn}}(k + 1 - t)} }}{{\sum\limits_{t = 0}^T {\sum\limits_m {\boldsymbol{\Delta u}_m^2\left( {k - t} \right)\boldsymbol{\theta} _m^2\left( {k - t} \right)} } }}} } \boldsymbol{\tilde W}_{mn}(k)\\
		&  + \frac{{k_2^2\sum\limits_{t = 1}^T {{{\left\| {{{\left[ {\boldsymbol{\Delta y}\left( {k + 1 - t} \right) - \boldsymbol{\Delta \hat y}\left( {k + 1 - t} \right)} \right]}^{\rm{T}}}} \right\|}^2}} \sum\limits_m {\sum\limits_n {\sum\limits_{t = 1}^T {\boldsymbol{\Delta u}_m^2\left( {k + 1 - t} \right)\boldsymbol{\theta} _{mn}^2(k + 1 - t)} } } }}{{{{\left[ {\sum\limits_{t = 0}^T {\sum\limits_m {\boldsymbol{\Delta u}_m^2\left( {k - t} \right)\boldsymbol{\theta} _m^2\left( {k - t} \right)} } } \right]}^2}}}\\
		\le  & \frac{{\left( {k_2^2 - 2{k_2}} \right)\sum\limits_{t = 1}^T {{{\left\| {{{\left[ {\boldsymbol{\Delta y}\left( {k + 1 - t} \right) - \boldsymbol{\Delta \hat y}\left( {k + 1 - t} \right)} \right]}^{\rm{T}}}} \right\|}^2}} }}{{\sum\limits_{t = 0}^T {\sum\limits_m {\boldsymbol{\Delta u}_m^2\left( {k - t} \right)\boldsymbol{\theta} _m^2\left( {k - t} \right)} } }}
	\end{aligned}
\end{equation}
It is obvious that ${V_2}(k + 1) - {V_2}(k)$ should be negative, and the positive number $k$ ranges from 0 to 2. 

For the convenience of readers, the detailed block diagram of the control method in this paper can be seen in Fig.2.

\begin{figure}[htbp]
	\centering
	\includegraphics[width=12cm]{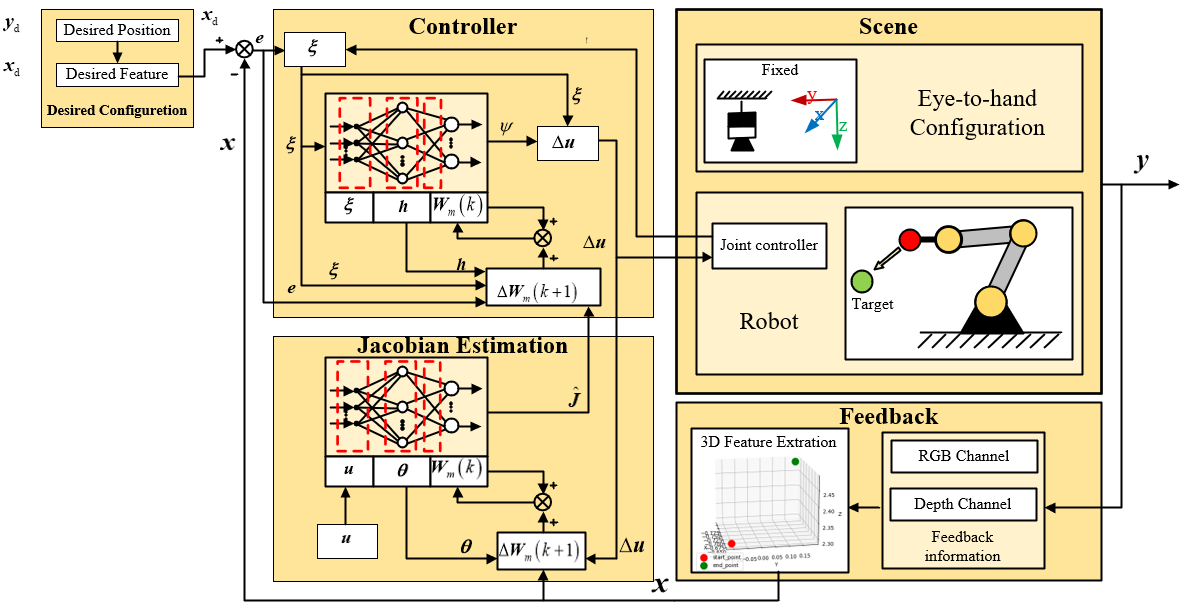}
	\caption{The block diagram of the control algorithm proposed in this paper.}
\end{figure}

\section{Experimental Results}
\noindent In this section, a series of experiments are undertaken to test the robustness and convergent performance of the control algorithm suggested in this research. Due to the importance of the estimate accuracy of the Jacobian matrix to the performance of the system, both estimation error and convergency error will be measured in order to investigate the characteristics of various estimation methods and controllers.

In CoppeliaSim, simulation experiments are conducted. A UR5 robot arm and a depth camera are utilized for this project. Furthermore, the depth camera is suspended above the robot arm. A red ball is attached to the end of the robot arm to facilitate image recognition. And the layout is depicted in Figure 3.

\begin{figure}[htbp]
	\centering
	\includegraphics[width=8cm]{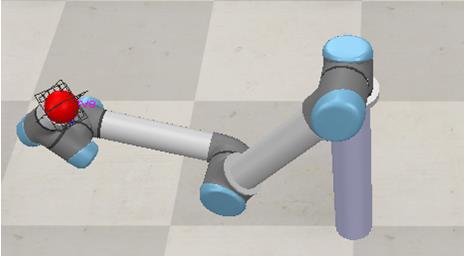}
	\caption{The overview of simulation environment in CoppeliaSim. The end-effector is bound with a red ball for easier image identification.}
\end{figure}

\subsection{Stationary Target Manipulation}
\noindent In this section, the velocity controller drives the UR5 robot to move the end-effector to the stationary required position. Three more control experiments are conducted here to demonstrate the performance of the algorithm provided in this research. One derivation from \cite{17} is RBF+PID, in which the Jacobian matrix is computed using RBFNN and the neural network is updated during the control process. It will be based on the same neural network as the estimator in our paper, allowing the influence of different controller and estimator update algorithms to be shown. The remaining two controllers, designated MPC and MFAC, are based on the UKF estimator. The specifics of their designs are as follows:

\begin{itemize}
	\item {Method1:RBF+PID scheme formula \cite {17}:}\\
		In this scheme, the estimator of the Jacobian matrix is constructed using a RBFNN, which is identical to the estimator in our paper except for the update algorithm. This scheme's controller, however, is designed solely on the basis of the decrease of the Lyapunov function; this is the primary distinction between this scheme and the proposed scheme. The specifics of its design are as follows: The RBFNN estimator from (40) to (41) is identical to the estimator in the proposed scheme, whereas the update algorithm in (42) is unique. The controller is displayed in (43).
		\begin{equation}
			{\boldsymbol{\theta} _{{{it}}}} = {{\rm{e}}^{{{ - }}\left[ \frac{{||\boldsymbol{u} - {\boldsymbol{u}_{{{it}}}}||}}{{{\boldsymbol{\delta} _{{{it}}}}}}\right] {^2}}}  = {{\rm{e}}^{{{ - }}\left[ \sqrt {\sum\limits_{{{j}} = 1}^{{l}} {{{(\frac{{{\boldsymbol{u}_{{j}}} - {\boldsymbol{u}_{{{itj}}}}}}{{{\boldsymbol{\delta} _{{{it}}}}}})}^2}} }\right]  {^2}}}
		\end{equation}  
		\begin{equation}
			{\hat {\boldsymbol{J}}_{{i}}} = {\boldsymbol{W}_{{i}}}{\boldsymbol{\theta} _{{i}}}(\boldsymbol{u}),{{i}} = 1,...,6
		\end{equation}
		\begin{equation}
			{\dot {\hat {\boldsymbol{W}}}}^\mathrm{T}_{{{ij}}} = {\dot {\boldsymbol{u}}_{{i}}}{\boldsymbol{\theta} _{{i}}}({{{n}}_2}{{\boldsymbol{e}}_{{j}}} + {n_3}{\boldsymbol{s}_j})
		\end{equation}
		\begin{equation}
			{\dot {\boldsymbol{u}} =  - }k{\left( {\hat {\boldsymbol{J}}} \right)^ + }\Delta \boldsymbol{x}
		\end{equation}  
		By deriving the Lyapunov function ${\rm{V}} = \frac{1}{2}{n_2}\Delta {\boldsymbol{x}^\mathrm{T}}\Delta \boldsymbol{x} + \frac{1}{2}\sum\limits_{{\rm{i}} = 1}^{\rm{n}} {\sum\limits_{{\rm{j}} = 1}^{\rm{l}} {\Delta {\boldsymbol{W}_{{\rm{ij}}}}\Delta \boldsymbol{W}_{{\rm{ij}}}^\mathrm{T}} } $, 
		it can be obtained that:
		\begin{equation}
			\begin{aligned}
				{\rm{\dot V}} =&{n_2}\Delta {\boldsymbol{x}^\mathrm{T}}\Delta \dot {\boldsymbol{x}} - \sum\limits_{{\rm{i}} = 1}^{\rm{n}} {\sum\limits_{{\rm{j}} = 1}^{\rm{l}} {\Delta {\boldsymbol{W}_{{\rm{ij}}}}{\dot {\hat {\boldsymbol{W}_{{\rm{ij}}}}}}^\mathrm{T}} } \\
				=& {n_2}\Delta {\boldsymbol{x}^\mathrm{T}}(\boldsymbol{e} + \hat {\boldsymbol{J}}\dot {\boldsymbol{u}}) - \sum\limits_{{\rm{i}} = 1}^{\rm{n}} {\sum\limits_{{\rm{j}} = 1}^{\rm{l}} {\Delta {\boldsymbol{W}_{{\rm{ij}}}}{{\dot {\boldsymbol{u}}}_{\rm{i}}}{\boldsymbol{\theta} _{\rm{i}}}({{\rm{n}}_2}\Delta {\boldsymbol{x}_{\rm{j}}} + {{\rm{n}}_3}{\boldsymbol{e}_{\rm{j}}})} } \\
				=&  - {\rm{k}}{n_2}\Delta {\boldsymbol{x}^\mathrm{T}}\Delta \boldsymbol{x} - {{\rm{n}}_3}{\boldsymbol{e}^\mathrm{T}}\boldsymbol{e} \le 0
			\end{aligned}
		\end{equation}  
	
	\item {Method2:UKF+MFAC scheme formula \cite {adaptive}:}\\
		In this scheme, the Jacobian matrix estimator is built using the Unscented Kalman Filter (UKF), which has both optimal initial estimation and control process performance. In this method, a model-free adaptive controller (MFAC) is utilized. Similarly, the performance index is created by combining system input and system error. By setting the performance index's derivative with respect to input to zero, the velocity command can be calculated. The specifics of its construction and the formulation of the UKF estimator can be found in  \cite {adaptive}.\\
		The model free adaptive controller is designed as:
		\begin{equation}
			{\boldsymbol{r}_k} = {\boldsymbol{r}_{k - 1}} + {\left( {\lambda {{\boldsymbol{E}}_q} + {\boldsymbol{\hat J}}_k^{\rm{T}}{{{\boldsymbol{\hat J}}}_k}} \right)^{ +}}{\boldsymbol{\hat J}}_k^{\rm{T}}{\boldsymbol{e}_{k - 1}}
		\end{equation}     
		where $\lambda $ is the weight that controls the magnitude of $\boldsymbol{\Delta r}_k$. Note that the model estimation algorithm (like UKF) exactly approximates the Jacobian matrix such that ${{\boldsymbol{J}}_k} = {{\boldsymbol{\hat J}}_k}$.
		
	\item {Method3:UKF+MPC scheme formula \cite {towards}:}\\
		The Jacobian matrix estimator is constructed using the UKF estimator in this scheme. A model predictive controller (MPC) is utilized here, and all states within the prediction horizon are considered. In the MPC controller, a criterion that takes into account both manipulation smoothness and system error is designed, and the velocity command can be derived by setting its derivative with respect to input to zero. The following are the specifics of the model predictive controller's design:
		\begin{equation}
			\begin{aligned}
				{\boldsymbol{r}_k} =  & {\boldsymbol{r}_{k - 1}} + {\left( {a{{{\boldsymbol{\hat J}}}_k} + {\boldsymbol{\hat J}}{{_k^{\rm{T}}}^ + }{\boldsymbol{Q}}} \right)^ + }(b - c){\boldsymbol{e}_k}\\
				a =  & {{\left( {{H^2}{\alpha ^ * }^H - 2b} \right)} \mathord{\left/
						{\vphantom {{\left( {{H^2}{\alpha ^ * }^H - 2b} \right)} {\ln {\alpha ^ * }}}} \right.
						\kern-\nulldelimiterspace} {\ln {\alpha ^ * }}}\\
				b =  & {{\left( {H{\alpha ^ * }^H\ln {\alpha ^ * } - {\alpha ^ * }^H + 1} \right)} \mathord{\left/
						{\vphantom {{\left( {H{\alpha ^ * }^H\ln {\alpha ^ * } - {\alpha ^ * }^H + 1} \right)} {{{\ln }^2}{\alpha ^ * }}}} \right.
						\kern-\nulldelimiterspace} {{{\ln }^2}{\alpha ^ * }}}\\
				c =  & {{\left( {H{{\left( {{\alpha ^ * }\beta } \right)}^H}\ln \left( {{\alpha ^ * }\beta } \right) - {{\left( {{\alpha ^ * }\beta } \right)}^H} + 1} \right)} \mathord{\left/
						{\vphantom {{\left( {H{{\left( {{\alpha ^ * }\beta } \right)}^H}\ln \left( {{\alpha ^ * }\beta } \right) - {{\left( {{\alpha ^ * }\beta } \right)}^H} + 1} \right)} {{{\ln }^2}\left( {{\alpha ^ * }\beta } \right)}}} \right.
						\kern-\nulldelimiterspace} {{{\ln }^2}\left( {{\alpha ^ * }\beta } \right)}}
			\end{aligned}
		\end{equation}
		where $H$ is the length of prediction horizon, $0 < \alpha^ *  \le 1$, $\beta  = \exp \left( { - \rho } \right)$, $\rho$ is a positive constant, and ${\boldsymbol{Q}}$ is a symmetric and positive definite matrix used to adjust the command.
		
\end{itemize}

We conducted four trials with a variety of stationary initial and target positions. The results of the trials are depicted in Fig.4. The convergency graphs demonstrate that the proposed control scheme significantly increases the convergency rate. In addition, if not properly trained, the use of neural networks to estimate Jacobian matrixes may negatively impact system performance. Frome Fig.4a and Fig.4d, it can be seen that both the proposed scheme and the RBF+PID scheme are badly influenced by the estimator model fault.

\begin{figure}[H]
	\centering
	\subfloat[]{
		\includegraphics[width=1.95 in]{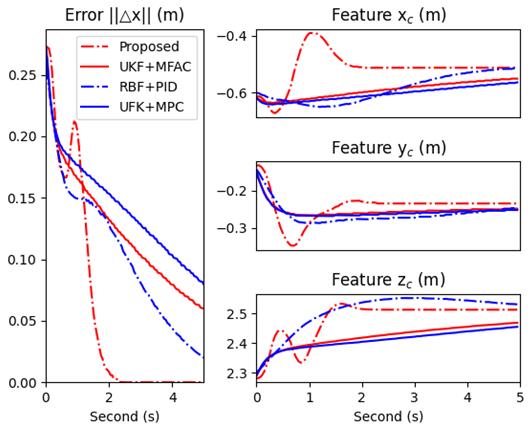}}
	\subfloat[]{
		\includegraphics[width=1.95 in]{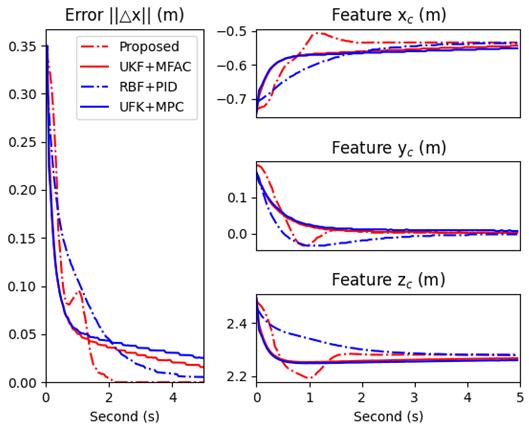}}\vspace{-0.3cm}
	\quad
	\subfloat[]{
		\includegraphics[width=1.95 in]{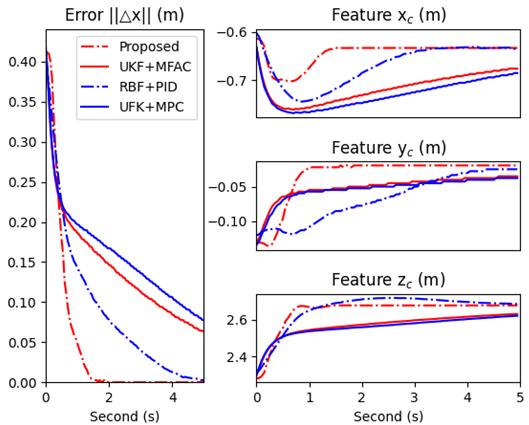}}
	\subfloat[]{
		\includegraphics[width=1.95 in]{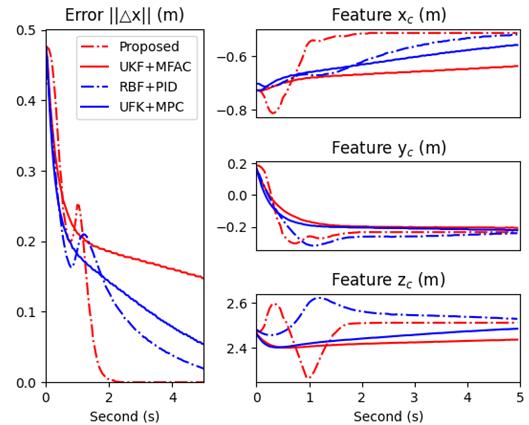}}\vspace{-0.3cm}
	\caption{Among four methods, the profiles of error of the four visual servoing experiments and value of image feature captured by depth camera. (a) EXP1. (b) EXP2. (c) EXP3. (d) EXP4 .}
\end{figure}

In total, four schemes are employed in the tests. The "RBF+PID" scheme designs its update algorithm and controller when considering both the system output error and the estimator error. The other two that have separate estimators and their controllers rely on the reliability of estimation are "UKF+MFAC" and "UKF+MPC". The performance of the MPC controller and the MFAC controller is comparable. Since their estimators are entirely based on system latest input-output information. Due to the combination of MFAC designing and "receding-horizon" designing from MPC, the proposed scheme performs better in terms of system error convergence. However, the overshoot of the proposed method is more obvious. Faster convergence seems to make it more susceptible to errors from the offline trained models (shown in Fig.4.a and Fig.4.d) compared with the "RBF+PID" scheme.

Fig.5 shows the 3-dimensional plot trajectory of the manipulator's end-effector from the perspective of a depth camera, to illustrate the trajectory of the end-effector for various controllers. 
In this three-dimensional locus diagram, the overshoot phenomenon of proposed scheme is more obvious. This is specifically reflected in the fact that its trajectory is likely to be a longer path compared with other three schemes.

\begin{figure}[H]
	\centering
	\subfloat[]{ 
		\includegraphics[width=1.95 in]{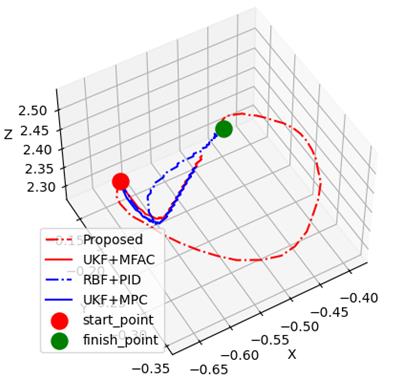}}
	\subfloat[]{
		\includegraphics[width=1.95 in]{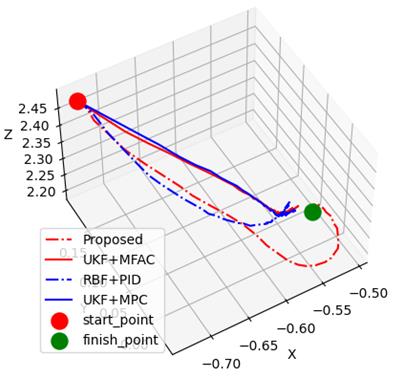}} \vspace{-0.3cm}
	\quad
	\subfloat[]{
		\includegraphics[width=1.95 in]{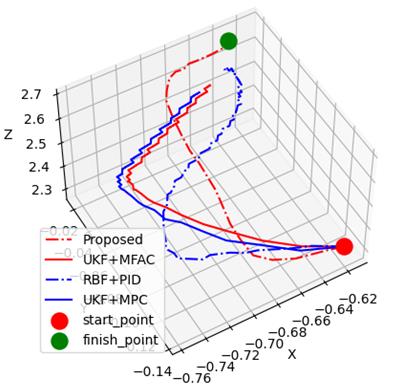}}
	\subfloat[]{
		\includegraphics[width=1.95 in]{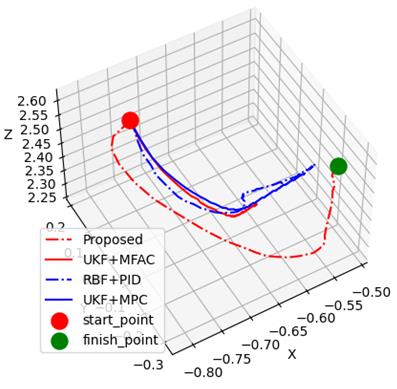}} \vspace{-0.3cm}
	\caption{The 3-dimensional plot of the end trajectory in the four visual servoing experiments which have a variety of different start and target positions with four control algorithms, and the current position is at the target position. (a) EXP1. (b) EXP2. (c) EXP3. (d) EXP4 .}
\end{figure}

\subsection{Trajectory tracking}
\noindent In this part, the proposed velocity controller drives the UR5 robot in order to drive the end-effector to track the trajectory designed before. We designed a circular trajectory in the camera coordinate system with a radius of 0.3 meters, divided into 200 discrete points, and let the end of the robotic arm track this trajectory. It is worth mentioning that the simulation experiment is still carried out in "CoppeliaSim", and the experimental environment is different from the environment where the training data is collected and our proposed update law for Jacobian estimator can help the estimation keep path with the true value.

The target trajectory is a horizontal circular trajectory in the camera coordinate system, so the target features of the x-axis and y-axis are sinusoidal curves with the same amplitude and period, while the feature of the z-axis are horizontal straight line. The three-dimensional characteristics of the target curve in the camera coordinate system, the three-dimensional characteristics of the feedback signal and the real-time error distance can be seen in Fig.6. In addition, in order to allow readers to have a more intuitive understanding of the control trajectory, the three-dimensional feedback characteristics in the camera coordinate system and the three-dimensional diagram of the design trajectory characteristics can be seen in Fig.7 and Fig.8.
It can be seen from Fig.6 that despite the large initial error, our controller is still able to quickly drive the end-effector close to the trajectory and then keep the end-effector on track to the target trajectory. It is noted that all the data used to train the offline model is collected via the DH model. Our update law enables the offline model to track the actual model, which means we can use a simplified model to save time and effort in collecting data.
\begin{figure}[htbp]
	\centering
	\includegraphics[width=7cm]{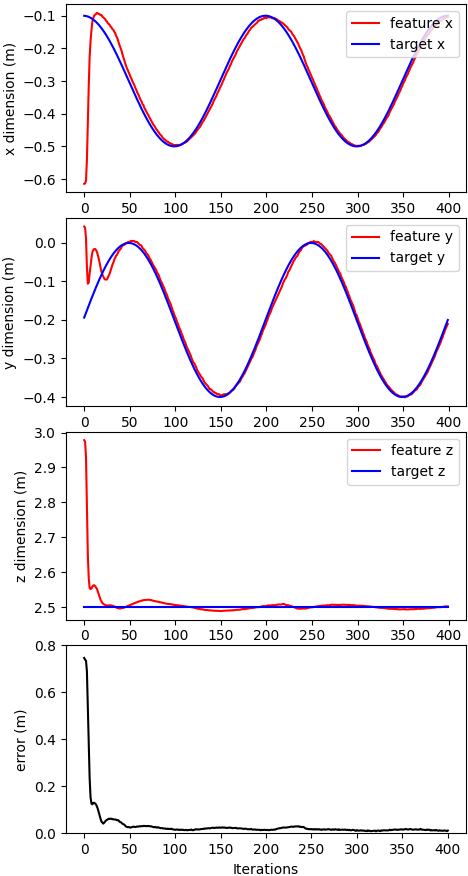}
	\caption{ the profiles of error of the trajectory tracking experiments and value of feedback/designed feature captured by depth camera.}
\end{figure}

\begin{figure}[htbp]
	\centering
	\includegraphics[width=8cm]{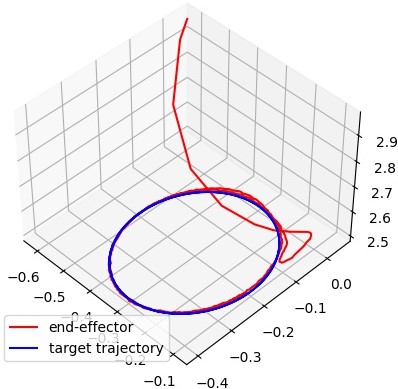}
	\caption{ the profiles of the 3-dimensional plot of the feedback and designed trajectory}
\end{figure}

\begin{figure}[H]
	\centering
	\subfloat[]{
		\includegraphics[width=1.6 in]{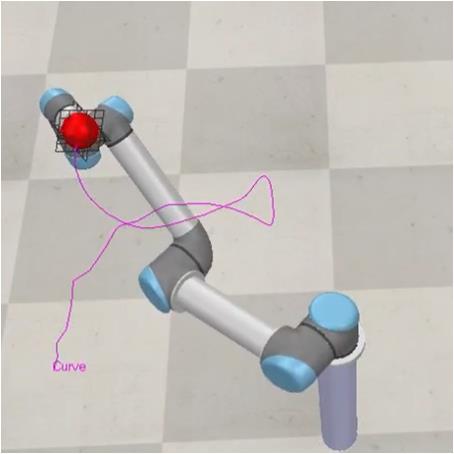}}
	\subfloat[]{
		\includegraphics[width=1.6 in]{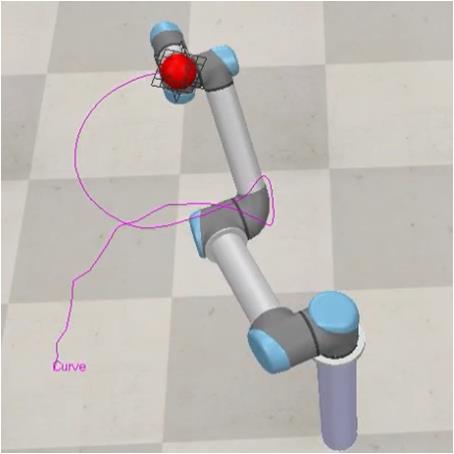}}
	\subfloat[]{
		\includegraphics[width=1.6 in]{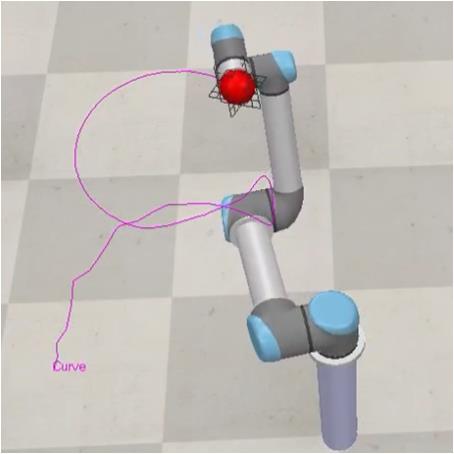}}
	\subfloat[]{
		\includegraphics[width=1.6 in]{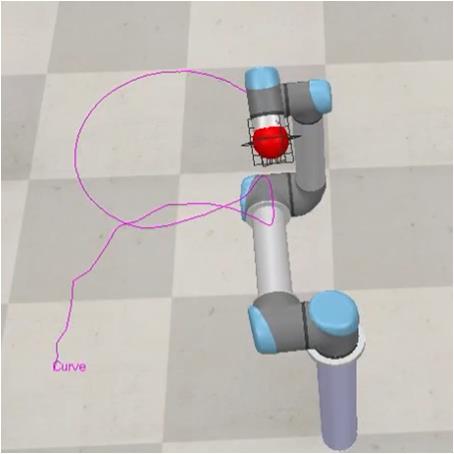}}\vspace{-0.3cm}
	\quad
	\subfloat[]{
		\includegraphics[width=1.6 in]{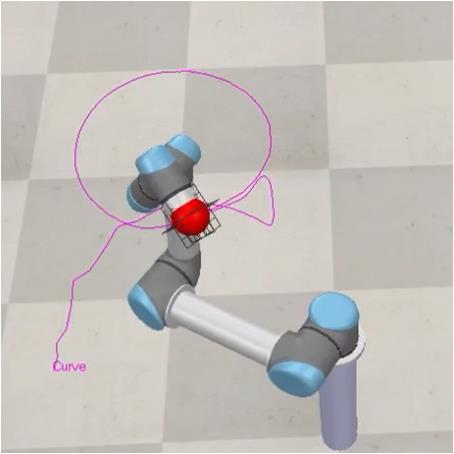}}
	\subfloat[]{
		\includegraphics[width=1.6 in]{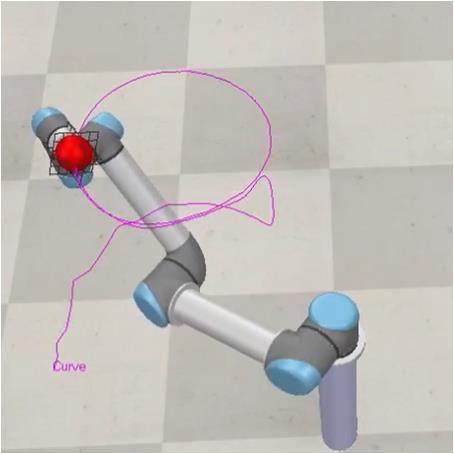}}
	\subfloat[]{
		\includegraphics[width=1.6 in]{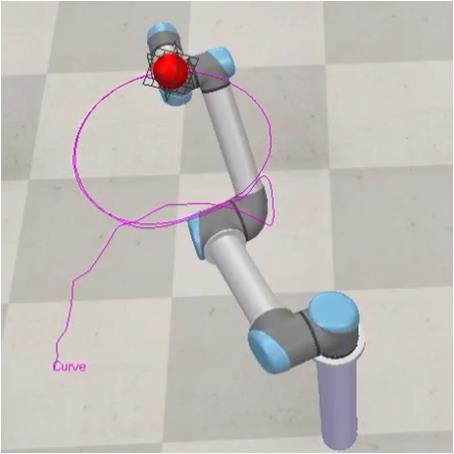}}
	\subfloat[]{
		\includegraphics[width=1.6 in]{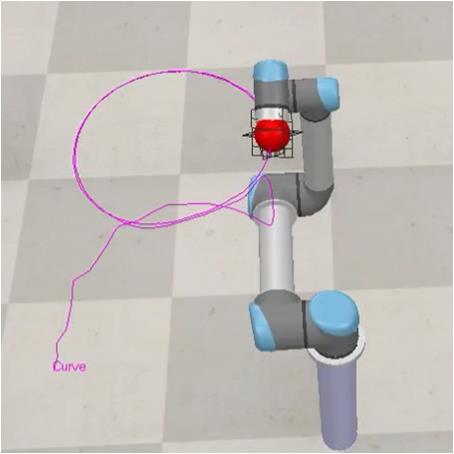}}\vspace{-0.3cm}
	\caption{The real-time trajectory diagram of the robot arm trajectory tracking control in the "CoppeliaSim" simulation software}
\end{figure}

\section{Conclusion}
\noindent In this work, we offer a MFAC technique that combines offline and online methods, with an adaptive neural network used to estimate the local hand-eye relationship. The proposed estimator performs better than the estimators that exclusively employ online methods because it can achieve a balance between performance and robustness by combining online and offline methods. Additionally, the suggested algorithm's converges more quickly with the "receding-horizon" designing. The adaptive update step size makes the controller and estimator numerically proven to converge to the true value. The "receding-horizon" designing inspired by MPC method help us achieve our goal of trajectory tracking.

The main innovation of this paper is that we propose a method for controller design by obtaining real-time mapping matrices. In addition, the structural characteristics of the combination of offline and online components make our control scheme not only have a better convergence performance, but also adapt to new environments and situations. The combination of the MFAC method and the "receding-horizon" structure in the MPC method makes our controller converge faster when targeting stationary targets, and its function is extended to track target trajectories.

Despite its performance in convergence speed and adaptation, the proposed method seems to suffer more from offline model errors and its overshoot phenomenon is more obvious. However, these problems can be alleviated by adjusting the parameters of ${k}$ and ${k_2}$ in the step size formula.

\addtolength{\textheight}{-1.50cm} 

\bibliographystyle{IEEEtran}
\bibliography{ref.bib}

\end{document}